\documentclass[letterpaper]{article}
\usepackage[preprint]{aaai2027}
\usepackage[hyphens]{url}
\usepackage{graphicx}
\usepackage{natbib}
\usepackage{caption}
\usepackage{amsmath}
\usepackage{amsfonts}
\usepackage{booktabs}
\usepackage{listings}

\newcommand{\tool}{VerClin}

\title{Refining and Aligning Clinical Reasoning with Health Guidelines via Formal Constraints}
\author {
    Xiaoyang Fan\textsuperscript{\rm 1},
    Yufan Cai\textsuperscript{\rm 1},
    Zhe Hou\textsuperscript{\rm 2},
    Jin Song Dong\textsuperscript{\rm 1}
}
\affiliations {
    \textsuperscript{\rm 1}National University of Singapore\\
    \textsuperscript{\rm 2}Griffith University\\
    x.fan@u.nus.edu, cai\_yufan@u.nus.edu, z.hou@griffith.edu.au, dcsdjs@nus.edu.sg
}

\begin{document}

\maketitle

\begin{abstract}
Clinical reasoning from incomplete or conflicting patient evidence must account for low-probability but high-risk concerns while remaining grounded in health guidelines. 
We introduce \tool{}, an explainable framework that extracts guideline conditions, exclusions, recommendations, and priorities into executable knowledge and represents patient conversations as scoped cases. 
A deterministic four-valued reasoner produces traceable guideline matches while preserving uncertainty and enforcing exclusions. 
Multiple specialty agents then screen matched guidelines for case applicability, and a consultation model synthesizes accepted guidance through specialty recommendations and a sentence-preserving formal-constraint channel. 
We present accepted guidance in ascending order of the sum of its priority and trigger-strength ranks, which can surface high-risk guidance with weaker triggers.
\tool{} outperforms direct generation and Top-5 RAG on HealthBench Hard and HealthBench Professional. On the LLMEval-Med subset, it outperforms direct generation and ties Top-5 RAG.
Component analyses further examine semantic closure, example-guided refinement, specialty screening, and constraint-conditioned delivery.
\end{abstract}

\section{Introduction}

Bridging natural-language understanding and disciplined clinical reasoning remains a central challenge for medical artificial intelligence, where evaluations span diverse clinical tasks and settings~\cite{ijcai2025p1169}.
Large language models (LLMs) can reason over medical problems with verifiable answers~\cite{chen-etal-2025-towards-medical}, but a fluent response alone does not show whether relevant guideline conditions were identified, exclusions were respected, or high-risk guidance was made salient.
MultifacetEval reports difficulty with answer rectification and verification~\cite{ijcai2024p737}, and an analysis of HealthBench documents limits in its clinical coverage~\cite{liu2025dissecting}.
These findings motivate explicit interfaces between authored guidance, patient-specific evidence, and natural-language response generation.
Such interfaces are especially important when evidence is incomplete or conflicting and when a \textbf{lower-probability} concern carries \textbf{high risk.}

We present \textbf{\tool{}}, a formal-constraint--driven framework for refining and aligning LLM-based clinical reasoning with health guidelines.
\tool{} compiles guideline triggers and exclusions into executable Rule IR, represents each conversation as scoped Case IR, and evaluates their correspondence with a deterministic four-valued reasoner.
Controlled semantic closure requests model judgments only for unresolved propositions that can still affect a guideline disposition, after which the deterministic reasoner recomputes the result.
Guideline priority is kept distinct from trigger strength and orders accepted guidance in the response context so that higher-risk material is presented earlier.
After formal matching, multiple specialty agents independently assess whether each matched guideline is applicable to the current case.
Here, \emph{multi-agent} denotes parallel specialty assessors and consultants rather than agents that generate independent diagnostic candidates for later optimization.
Guidance accepted by at least one assigned specialty is supplied to a constraint-conditioned consultation model through both specialty recommendations and a sentence-preserving formal-constraint channel.
The consultation stage synthesizes these inputs into a response; it is prompt-conditioned generation rather than a formal optimizer or an output-verification procedure.

We evaluate \tool{} on HealthBench Hard, HealthBench Professional, and the medical-reasoning and medical-safety-and-ethics subset of LLMEval-Med.
\tool{} outperforms the reported direct-generation and Top-5 RAG baselines on both HealthBench benchmarks and matches Top-5 RAG on the selected LLMEval-Med subset.
Component ablations study semantic closure, example-guided refinement, and the formal-constraint channel, while a separate analysis characterizes specialty screening.

\paragraph{Contributions.}
\begin{itemize}
    \item We introduce an executable guideline representation and a scoped Case IR connected by formal reasoning, hard exclusion, and controlled semantic closure.
    \item We develop example-guided refinement from auditable formal failure traces while preserving source-anchored guideline conditions and explicit intermediate artifacts.
    \item We define a multi-agent specialty-screening and constraint-conditioned consultation workflow in which guideline-grounded content remains traceable to its source, trigger, and case evidence.
    \item We evaluate final response quality and the behavior of the refinement, screening, and delivery components across three medical benchmarks.
\end{itemize}

\section{Motivating Example}
\label{sec:motivating_example}

\paragraph{Illustrative clinical scenario.}
Consider a 52-year-old man with intermittent chest tightness lasting four days.
The symptoms occur mainly during exertion, such as climbing stairs, and resolve with rest.
He reports mild shortness of breath but denies nausea, diaphoresis, or syncope.
His vital signs are stable, the ECG shows non-specific ST changes, and the initial troponin is normal.
He recently completed a 13-hour flight and reports mild soreness in his left calf.
He asks whether he can be safely discharged if the test results ``look okay.''
This constructed example illustrates the intended workflow and is not presented as an evaluated benchmark output.

No single finding is strongly abnormal, yet the combination contains weak evidence for low-probability but high-risk conditions.
A likelihood-centered GPT-5.2 response could emphasize the reassuring findings and reason as follows:

\begin{quote}
\small
``The discomfort is exertional and relieved by rest, but the initial troponin is normal and the presentation lacks several classic ischemic features.
Acute coronary syndrome therefore appears less likely, and discharge with outpatient follow-up may be reasonable if the remaining evaluation is reassuring.''
\end{quote}

\begin{table*}[t]
\centering
\begingroup
\small
\setlength{\tabcolsep}{4pt}
\renewcommand{\arraystretch}{1.18}
\begin{tabular}{p{0.16\textwidth}p{0.34\textwidth}p{0.40\textwidth}}
\toprule
\multicolumn{1}{c}{\textbf{Aspect}} & \multicolumn{1}{c}{\textbf{GPT-5.2 Default}} & \multicolumn{1}{c}{\textbf{\tool{}}} \\
\midrule
\multicolumn{1}{c}{\textbf{Evidence use}} &
Normal troponin and stable vital signs dominate the response, while the travel history, dyspnea, and calf discomfort receive little attention. &
Case IR keeps the exertional symptoms, dyspnea, travel history, calf discomfort, reassuring tests, subjects, and temporal qualifiers in separate evidence-bearing claims. \\
\addlinespace[2pt]
\multicolumn{1}{c}{\textbf{Guideline match}} &
No explicit record identifies which guideline conditions were considered or remained unresolved. &
Executable triggers match guidance concerning ischemic and thromboembolic evaluation, while hard exclusions remain available to prevent incompatible guidance. \\
\addlinespace[2pt]
\multicolumn{1}{c}{\textbf{Specialty review}} &
The response is produced from a single likelihood-oriented view. &
Cardiac and thromboembolic assessors independently review whether the formally matched guidance is applicable and provide case-specific recommendations. \\
\addlinespace[2pt]
\multicolumn{1}{c}{\textbf{Priority}} &
Likelihood-first organization favors reassurance. &
Higher-risk guidance is ordered before lower-priority material so that pulmonary embolism and acute coronary syndrome receive explicit attention despite uncertain probability. \\
\addlinespace[2pt]
\multicolumn{1}{c}{\textbf{Response}} &
Discharge with outpatient follow-up is presented as plausible. &
Constraint-conditioned consultation recommends completing urgent ischemic and thromboembolic assessment and advises against discharge until that evaluation is complete. \\
\bottomrule
\end{tabular}
\endgroup
\caption{Illustrative comparison of the baseline and the \tool{} workflow on the same constructed scenario.}
\label{tab:case_study}
\end{table*}

Table~\ref{tab:case_study} contrasts likelihood-centered generation with the operations that \tool{} makes inspectable.
The example does not claim that the listed conditions establish either diagnosis.
Instead, it shows why an initially reassuring result should not erase other case evidence before relevant guidance has been considered.

\paragraph{Why executable guideline matching is required.}
Guidelines commonly combine multiple findings, qualifiers, alternatives, and exclusions.
Representing these conditions as executable Rule IR makes it possible to distinguish a proved trigger, a contradicted condition, conflicting evidence, and information that is simply absent.
The distinction matters here because a normal initial troponin does not logically negate the travel history, dyspnea, calf discomfort, or their relevance to a separate thromboembolic guideline.
Example-guided refinement addresses recurring representation failures through source- and evidence-anchored corrections, but each final match must still be derived by the reasoner from an authored condition.

\paragraph{Why priority is required.}
Clinical probability and clinical consequence are not interchangeable.
A guideline labelled with higher risk priority is placed earlier in the consultation context, independently of whether its trigger was strong, moderate, or weak.
This ordering is designed to help clinicians notice lower-probability concerns whose omission could carry high cost, without converting priority into proof of diagnosis or overriding hard exclusions.

\paragraph{Why specialty screening is required.}
Formal matching operates conservatively over the available catalogue, so some matched guidance can remain contextually inappropriate even when a trigger path is supported.
Multiple specialty assessors therefore review the same formally matched guideline from their assigned clinical perspectives and label it applicable, partially applicable, or not applicable.
A guideline accepted by at least one assigned specialty is retained for response construction, while a guideline rejected by every assigned specialty remains in the audit artifacts but does not ground the response.

\paragraph{Why constraint-conditioned consultation is required.}
Specialty recommendations can overlap, qualify one another, or compress details from the original guidance.
\tool{} therefore gives the consultation model both the specialty recommendations and a sentence-preserving formal-constraint channel containing accepted guideline text, proved triggers, evidence, and priority.
The resulting response is generated under these instructions rather than solved by a cross-agent symbolic optimizer, and no post-generation checker is claimed.
The guideline-grounded portion nevertheless remains inspectable through the accepted guideline, its screening assessments, its proof trace, and its source evidence.

\section{Approach}

\begin{figure*}
    \centering
    \includegraphics[width=0.7\linewidth]{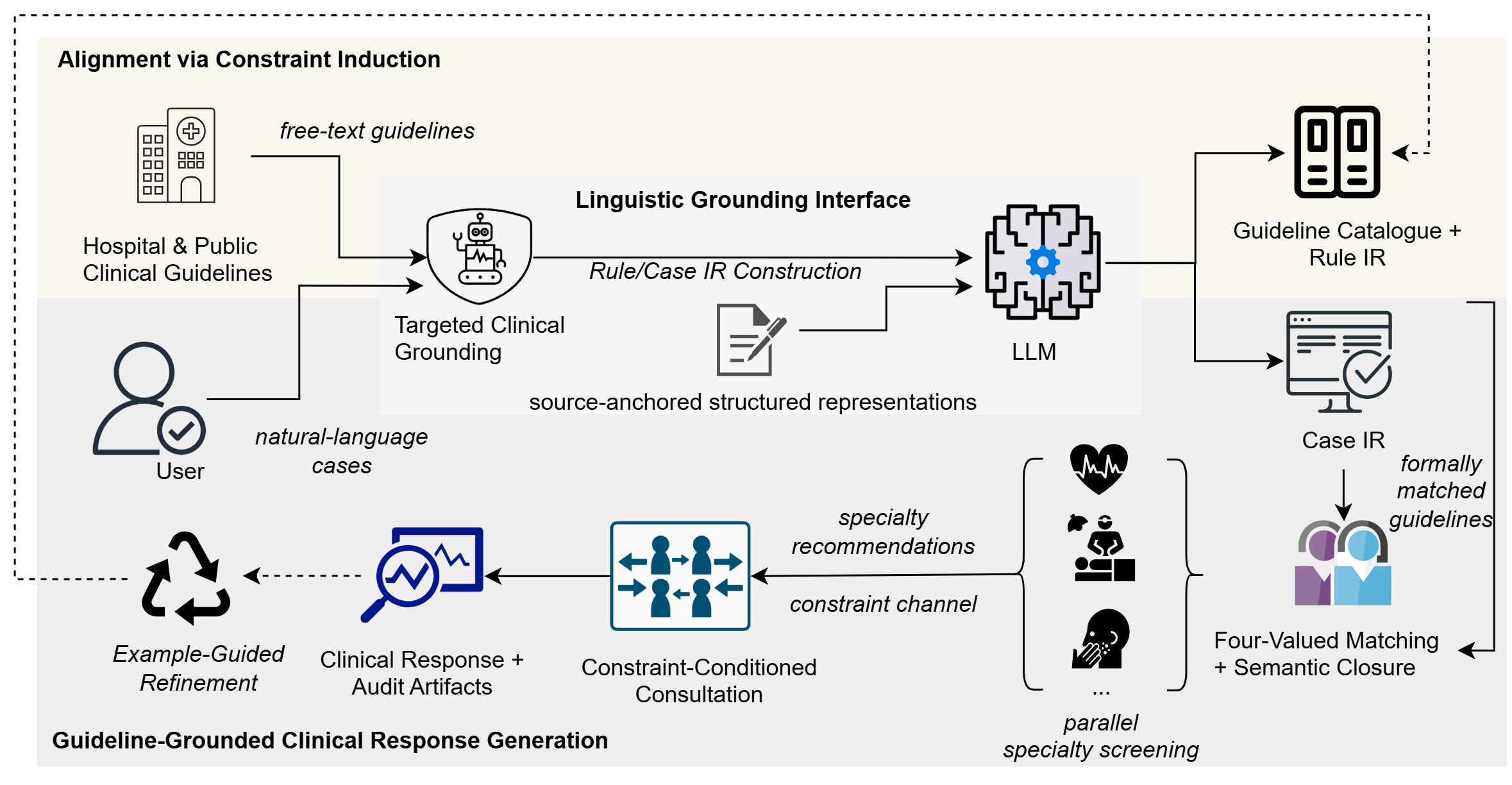}
    \caption{Overview of \tool{} from guideline and case representation through formal matching, multi-agent specialty screening, and constraint-conditioned consultation.}
    \label{fig:overview}
\end{figure*}

Figure~\ref{fig:overview} summarizes the end-to-end architecture of \tool{}, which connects open-ended medical language with source-anchored guideline conditions through a staged and inspectable interface.
Given a guideline catalogue and a free-text case, the framework constructs Rule and Case IR, evaluates their correspondence under open-world uncertainty, asks for bounded semantic judgments only where they can change a disposition, screens formally matched guidance through multiple specialty agents, and generates a response conditioned on the accepted guidance.
The formal reasoner controls logical composition, trigger strength, and hard exclusion, whereas specialty screening and final consultation remain model-based decisions.

\subsection{Alignment via Constraint Induction}

Each guideline $g$ in catalogue $\mathcal{G}$ contains source text $y_g$, a specialty set $S_g$, a clinical priority $P_g$, strong, moderate, and weak trigger sets $T_g^s,T_g^m,T_g^w$, and a hard-exclusion set $E_g$.
An advanced LLM induces the trigger and exclusion programs from authored guideline text and separately classifies priority and one or more specialties.
The trigger tiers encode the strength of case-to-guideline matching conditions, whereas $P_g\in\{\textsc{critical},\textsc{high},\textsc{medium},\textsc{low}\}$ records the risk importance of presenting the guidance after it has been accepted.
These fields therefore do not define a global ordering over candidate diagnoses.

\paragraph{Executable Rule IR.}
Each authored trigger or exclusion is compiled into a source-anchored logical program over the operators \textsc{all}, \textsc{any}, and \textsc{not}.
After structural validation and negation normalization, a condition has bounded disjunctive normal form:
\begin{equation}
c=\bigvee_i\bigwedge_j p_{ij}.
\end{equation}
Each proposition $p_{ij}$ is either a typed atom or a self-contained residual semantic proposition.
An atom records subject role, predicate, object text and linked concept candidates, assertion, temporality, comparison, value, unit, match mode, and source evidence.
Residual propositions preserve authored conditions that cannot be represented faithfully by the controlled vocabulary rather than forcing them into an overly broad symbolic approximation.

\paragraph{Example-guided refinement.}
Initial Rule IR and local clinical semantics are refined through audited formal failure traces.
The audit identifies unsupported concepts, directional relations, extraction gaps, or compiled conditions and records source- and evidence-supported semantic or rule-condition patches for replay.
Patches are localized and reusable, but the procedure does not claim to compute a formally minimal revision.
The resulting refinement pack is frozen before final answer generation and scoring, and no patch directly prescribes a case--guideline selection.

\subsection{Linguistic Grounding Interface}
\label{subsec:grounding}

Given a free-text case $x$, an advanced LLM constructs scoped Case IR:
$
\mathcal{F}(x)=\{\,\langle q,s,p,o,z,a,\tau,\nu,e\rangle\,\}.
$
Here, $q$ is a query scope, $s$ a clinical or conversational subject, $p$ a predicate, $o$ the object text, $z$ its concept candidates and linking audit, $a$ the assertion status, $\tau$ temporality, $\nu$ an optional comparator--value--unit tuple, and $e$ exact evidence from the input.
Query scopes separate different patients, questions, and requested tasks, preventing unrelated evidence from being combined into one proof.
The extractor records supported propositions and residual semantics.

\paragraph{Targeted clinical grounding.}
Only mentions requiring terminology support are linked to local registry entries and targeted medical concept candidates.
Exact verified mappings and directionally valid concept relations may support proof.
Ambiguous, broader, narrower, or provisional relations remain inspectable but evaluate conservatively as unknown unless a guarded local implication applies.

\paragraph{Open-world uncertainty.}
Every proposition is evaluated in the four-valued set
\begin{equation}
\mathbb{V}=\{\textsc{entailed},\textsc{contradicted},\textsc{both},\textsc{unknown}\}.
\end{equation}
Negation swaps \textsc{entailed} and \textsc{contradicted} while leaving \textsc{both} and \textsc{unknown} unchanged.
For conjunction, \textsc{contradicted} dominates, followed by \textsc{both}; the result is \textsc{entailed} only when every operand is entailed and is otherwise \textsc{unknown}.
For disjunction, \textsc{entailed} dominates, followed by \textsc{both}; the result is \textsc{contradicted} only when every operand is contradicted and is otherwise \textsc{unknown}.

\subsection{Formal Matching and Semantic Closure}

The deterministic reasoner matches Rule IR atoms against compatible Case IR claims while jointly respecting query scope, subject identity, predicate direction, assertion, temporality, quantities, and concept relations.
It evaluates the full dataset-specific catalogue symbolically.
Hybrid lexical and embedding retrieval only nominates unresolved residual propositions for possible semantic closure and cannot prevent a symbolic proof.

For a trigger tier $r\in\{s,m,w\}$, the attained match strength is
\begin{equation}
\begin{aligned}
\operatorname{strength}(g,x)
&=\max_{\succ}\{\,r:\exists c\in T_g^r,\\
&\qquad \operatorname{val}(c,x)=\textsc{entailed}\,\},
\end{aligned}
\end{equation}
where $s\succ m\succ w$.
An entailed hard exclusion overrides a proved trigger.
If an exclusion evaluates to \textsc{both} while a trigger is proved, the guideline remains unresolved; an unknown exclusion does not erase a proved trigger.
The reasoner thus records each guideline as formally matched at its strongest proved tier, hard-excluded, unresolved, or unmatched.

Semantic closure visits consequential unknowns in strong-to-moderate-to-weak order through a bounded frontier.
An advanced LLM assigns the same four-valued label using only indexed case evidence, and local validation checks label--evidence consistency, quoted evidence indices, and reason codes.
Invalid or unsupported judgments degrade to \textsc{unknown}.
Validated judgments are replayed through the deterministic reasoner until no consequential jobs remain or the configured pass limit is reached.

\paragraph{Risk-sensitive presentation order.}
Priority and trigger strength are combined only after formal matching to order accepted guidance for consultation.
Let $\rho(P_g)\in\{0,1,2,3\}$ rank critical through low priority and let $\sigma(r)\in\{0,1,2\}$ rank strong through weak triggers.
The presentation tier is
\begin{equation}
d(g,x)=1+\rho(P_g)+\sigma(\operatorname{strength}(g,x)),
\end{equation}
with lower values placed earlier and emphasized more strongly in the consultation context.
This mechanism is designed to surface high-risk guidance even when its trigger tier is weaker, but it is neither a diagnostic probability nor a formally verified safety guarantee.

\subsection{Multi-Agent Specialty Screening}

Clinical roles assign responsibility after formal matching rather than pruning the catalogue before reasoning.
A soft route proposes relevant specialties for the case, while the declared specialties of a matched guideline are restored when they fall outside that route.
Each specialty agent receives the case, guideline text, proved trigger, and evidence and returns an applicability label together with a concise recommendation.
The labels are \textsc{applicable}, \textsc{partially applicable}, and \textsc{not applicable}.
A formally matched guideline is accepted for response construction when at least one assigned specialty returns either of the first two labels, and a guideline rejected by every assigned specialty remains in the audit artifacts but is withheld from the response inputs.
The formal disposition and the screening decision are stored separately so that a model-based applicability judgment is not presented as part of the symbolic proof.

\subsection{Constraint-Conditioned Consultation}

Consultation receives the accepted specialty recommendations and a separate sentence-preserving formal-constraint channel.
For each accepted guideline, the channel contains the original guideline text, its sentence-level authored conclusions, proved triggers, case evidence, priority, trigger strength, and presentation tier.
The final GPT-5.2 model is instructed to synthesize these inputs into a coherent response and to include the accepted guidance, but no constrained decoder, output checker, or retry loop enforces this instruction.
Accordingly, \tool{} does not claim that generation is a formal realization layer or that every output statement follows from a proof.
The narrower traceability claim is that guideline-grounded content can be inspected through its accepted guideline, specialty assessments, formal match trace, source condition, and case evidence.

\section{Experiments}
\label{sec:experiments}

We have the following research questions:
\begin{itemize}
    \item \textbf{RQ1 (Answer quality):} Does \tool{} improve medical responses over direct generation and text-retrieval baselines?
    \item \textbf{RQ2 (Components):} What do semantic closure, example-guided refinement, and the formal-constraint channel contribute?
    \item \textbf{RQ3 (Guideline applicability):} How does multi-agent specialty screening assess and filter formally matched guidelines across benchmarks?
\end{itemize}


\paragraph{Datasets.}
We use three complementary medical benchmarks (Table~\ref{tab:v2_datasets}).
\textbf{HealthBench Hard} (HB-Hard) evaluates challenging healthcare conversations \cite{arora2025}, whereas \textbf{HealthBench Professional} (HB-Pro) evaluates clinician-chat tasks using a length-adjusted rubric score \cite{hicks2026healthbenchpro}.
\textbf{LLMEval-Med} spans medical knowledge (MK), medical reasoning (MR), medical language understanding (MLU), medical safety and ethics (MSE), and medical text generation (MTG), from which we retain the MR and MSE cases for the task-aligned evaluation described below \cite{zhang-etal-2025-llmeval}.
Following LLMEval-Med, we treat ratings of at least 4 on the 0--5 scale as usable. Overall performance (OP) is the fraction of usable responses across retained cases.

\begin{table}[t]
\centering
{
\setlength{\tabcolsep}{4pt}
\begin{tabular}{lrl}
\toprule
\textbf{Dataset} & \textbf{Cases} & \textbf{Primary metric} \\
\midrule
HB-Hard & 1,000 & Rubric score \\
HB-Pro & 525 & Length-adjusted score \\
LLMEval (MR+MSE) & 211 & Usability \\
\bottomrule
\end{tabular}
}
\caption{Reported evaluation sets and primary metrics.}
\label{tab:v2_datasets}
\end{table}


\paragraph{Baselines and main configuration.}
We compare \tool{} with GPT-4.1 direct generation, GPT-5.2 with default or high reasoning, and GPT-5.2 with Top-5 RAG.
The guidelines are extracted from the rubrics of the ordinary Healthbench dataset, excluding the test dataset to avoid possible data leakage.
The correctness of the Rule IR is verified by the original Healthbench dataset.
RAG and \tool{} access the same source-only catalogue and embedding infrastructure, but RAG only supplies retrieved text to the answer model; it is therefore an end-to-end retrieval comparator, not a control that isolates any single \tool{} component.
\tool{} constructs Case IR, retrieves semantic candidates, executes formal matching and controlled semantic closure, screens matched guidelines, and performs constraint-conditioned consultation.
%
HB-Pro and LLMEval-Med use the completed full-catalogue configuration with soft specialty routing, whereas HB-Hard retains an earlier completed configuration with specialty projection.
A replay of the later HB-Hard selector was stopped before submission because it projected roughly 36,000 additional semantic judgments, so the three datasets do not share an identical configuration.
%
Each system is generated and scored once under a fixed configuration.
%
Each run matches a case against the complete dataset-specific catalogue induced from benchmark-supplied guideline and rubric material.
Rule and Case IR, the terminology registry, semantic patches, retrieval indices, and catalogues are frozen unless an intervention directly changes them.
The refinement pack was built from target-benchmark failure traces and frozen before final scoring; on LLMEval-Med, it changes only external IR and semantic resources, never the evaluated LLM, so the results measure fitted benchmark performance rather than zero-shot generalization.

\subsection{Answer Quality (RQ1)}

\begin{table}[t]
\centering
{
\setlength{\tabcolsep}{3.5pt}
\begin{tabular}{lrrr}
\toprule
\textbf{System} & \textbf{HB-Hard} & \textbf{HB-Pro} & \textbf{LLMEval} \\
\midrule
GPT-4.1 & 13.47 & 26.96 & 60.19 \\
GPT-5.2 Default & 37.64 & 41.17 & 79.62 \\
GPT-5.2 High & 40.35 & 46.92 & 85.31 \\
Top-5 RAG & 45.71 & 69.89 & \textbf{90.52} \\
\tool{} & \textbf{50.44} & \textbf{74.38} & \textbf{90.52} \\
\bottomrule
\end{tabular}
}
\caption{Overall performance in percentage points, using the rubric score for HB-Hard, the length-adjusted score for HB-Pro, and micro-averaged usability on the LLMEval-Med subset.}
\label{tab:v2_rq1_overall}
\end{table}

\begin{figure}[t]
\centering
\includegraphics[width=0.8\columnwidth]{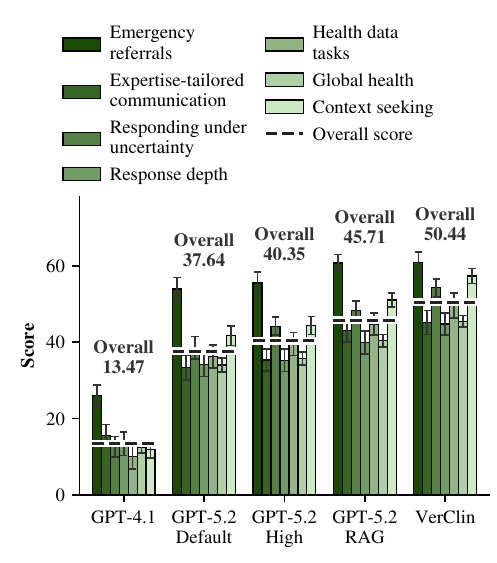}
\caption{HB-Hard scores by clinical theme, with bars for subgroup means, dashed lines for overall scores, and whiskers for one bootstrap standard deviation.}
\label{fig:v2_hb_hard_themes}
\end{figure}

\begin{figure}[t]
\centering
\includegraphics[width=0.8\columnwidth]{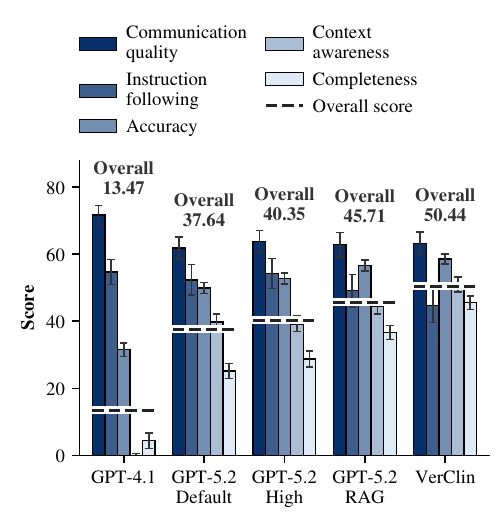}
\caption{HB-Hard scores by evaluation axis, with bars for axis means, dashed lines for overall scores, and whiskers for one bootstrap standard deviation.}
\label{fig:v2_hb_hard_axes}
\end{figure}

\paragraph{HealthBench Hard.}
\tool{} achieves an overall HB-Hard rubric score of 50.44, a statistically significant improvement of 4.73 percentage points over the Top-5 RAG score of 45.71.
Figures~\ref{fig:v2_hb_hard_themes} and~\ref{fig:v2_hb_hard_axes} show that the overall improvement is distributed across clinical themes and rubric axes rather than arising from a single category.

\begin{figure}[t]
\centering
\includegraphics[width=0.8\columnwidth]{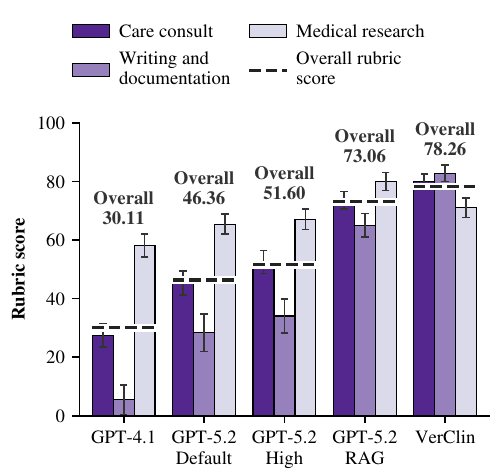}
\caption{Unadjusted HB-Pro scores by use case before length adjustment, with bars for subgroup means, dashed lines for overall scores, and whiskers for one bootstrap standard deviation.}
\label{fig:v2_hb_pro_use_case}
\end{figure}

\begin{figure}[t]
\centering
\includegraphics[width=0.8\columnwidth]{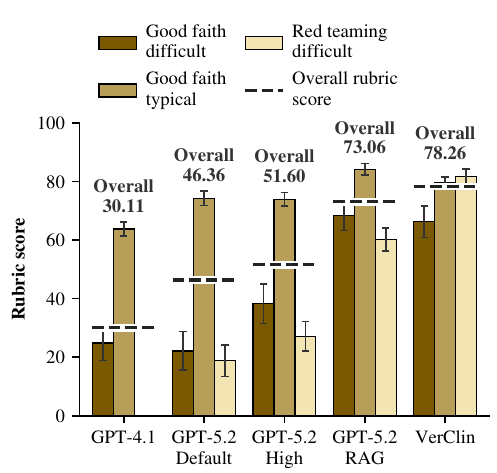}
\caption{Unadjusted HB-Pro scores for good-faith difficult, good-faith typical, and red-teaming difficult cases before length adjustment, with whiskers for one bootstrap standard deviation.}
\label{fig:v2_hb_pro_dataset_slice}
\end{figure}

\paragraph{HealthBench Professional.}
\tool{} also leads on HB-Pro; the aggregate is length-adjusted, whereas the category plots report the underlying rubric score.
\tool{} gains most in care consultation, documentation, and adversarially difficult cases, but trails RAG in medical research and both good-faith slices.
These unadjusted breakdowns localize the aggregate difference but are not numerically comparable with the length-adjusted primary score.

\begin{table}[t]
\centering
{
\setlength{\tabcolsep}{4.5pt}
\begin{tabular}{lrrr}
\toprule
\textbf{System} & \textbf{MR} & \textbf{MSE} & \textbf{OP} \\
\midrule
GPT-4.1 & 63.73 & 56.88 & 60.19 \\
GPT-5.2 Default & 70.59 & 88.07 & 79.62 \\
GPT-5.2 High & 80.39 & 89.91 & 85.31 \\
Top-5 RAG & \textbf{88.24} & 92.66 & \textbf{90.52} \\
\tool{} & 86.27 & \textbf{94.50} & \textbf{90.52} \\
\bottomrule
\end{tabular}
}
\caption{Usability on LLMEval-Med medical reasoning (MR; 102 cases) and medical safety and ethics (MSE; 109 cases), with OP as the micro-average over all 211 cases; bold denotes the best result in each column.}
\label{tab:v2_llmeval_subset}
\end{table}

\paragraph{LLMEval-Med reasoning-and-safety subset.}
We report only MR and MSE because they are the categories most directly aligned with guideline-conditioned reasoning and risk-sensitive response behavior.
We exclude MK and MLU because they primarily test factual recall and language comprehension rather than guideline-conditioned reasoning. We exclude MTG because its published results use a separate five-dimensional human evaluation~\cite{zhang-etal-2025-llmeval}.
This construct-aligned restriction narrows the empirical claim to reasoning and safety tasks and must not be interpreted as a benchmark-wide LLMEval-Med result.
\tool{} ties Top-5 RAG in overall usability, with RAG ahead on MR and \tool{} ahead on MSE; this task split does not identify a causal mechanism.
%
We define a comparative failure as a case on which \tool{} scores below Top-5 RAG, not as an independently verified clinical error.
Positive HealthBench means conceal case-level wins and losses, while the LLMEval tie contains a small, symmetric set of disagreements.
On HB-Hard, instruction following is the only axis on which \tool{} trails RAG, while completeness shows its largest advantage, suggesting a trade-off between coverage and conversational directness.
On HB-Pro, failures cluster in medical research and good-faith cases, whereas the largest gains occur on the adversarial slice; length adjustment narrows but does not reverse the unadjusted advantage.
On LLMEval, opposing MR and MSE differences cancel at the subset level.
Two potential mechanisms can produce these failures.
Weak case grounding can leave a valid rule unresolved; RQ2 characterizes this mechanism without causally labeling individual losses, while induced-rule errors remain outside the present evaluation.
Independent clinical adjudication remains necessary to assign responsibility for a particular mismatch.

\subsection{Component Analysis (RQ2)}

We conduct controlled ablations on HB-Pro, with each intervention paired against a content-matched full output from the same code and artifact configuration.
The interventions are defined as follows:
\begin{itemize}
    \item \textbf{No formal-constraint channel} keeps the same specialty-screening verdicts and response generator but removes the independent sentence-preserving constraint input to final consultation.
    \item \textbf{No example-guided refinement} removes the frozen local corrections learned from auditable formal failure traces.
    \item \textbf{No semantic closure} retains Rule and Case IR, concept linking, local typed semantics, deterministic matching, and hard exclusion but supplies no LLM semantic judgments for consequential unknown conditions.
\end{itemize}
Within the current-code group, removing semantic closure hurts more than removing refinement and leaves many matched guidelines unresolved, supporting its role at the language--logic boundary.
In the separate frozen answer-stage pair, removing the sentence-preserving channel also degrades performance.
Because the groups use different full references, their effect sizes are not ranked against one another.
A narrower diagnostic removes only the typed semantic core while preserving the remaining formal pipeline.
Its 0.54-point improvement ($p=.023$) shows no net benefit for the current typed core on HB-Pro, not evidence against structured grounding as a whole.

\begin{table}[t]
\centering
{
\small
\setlength{\tabcolsep}{2.5pt}
\begin{tabular}{lrr}
\toprule
\textbf{System} & \textbf{Score} & \textbf{$\Delta$ [95\% CI]} \\
\midrule
Frozen full \tool{} & 74.38 & -- \\
No constraint channel & 68.70 & $5.68\ [3.02,8.41]$ \\
\addlinespace[2pt]
Current-code full \tool{} & 74.20 & -- \\
No refinement & 69.45 & $4.76\ [2.19,7.52]$ \\
No semantic closure & 65.56 & $8.65\ [5.13,12.27]$ \\
\bottomrule
\end{tabular}
}
\caption{Single-column HB-Pro ablations with their paired full references, where $\Delta$ is reference minus ablation in percentage points and all three comparisons have $p<.001$.
Comparisons are made only within the frozen answer-stage pair or the current-code selector group.}
\label{tab:v2_ablation}
\end{table}

\subsection{Guideline Applicability Screening (RQ3)}

Each assigned specialty rates a formally matched pair as applicable, partially applicable, or not applicable, and \tool{} retains it when at least one specialty accepts it.
We report complete HB-Hard and HB-Pro results and the same LLMEval MR+MSE subset as RQ1; these model-produced labels characterize screening behavior, not clinical accuracy.
Specialty screening retains about nine tenths of matched pairs and modestly reduces the mean guidelines per case across all three benchmarks.
Rejection ranges from 13.0\% to 20.1\% for cross-case guidance but never exceeds 1.1\% for case-authored guidance, consistent with contextual filtering but not a substitute for clinical annotation.

\begin{figure}[t]
\centering
\includegraphics[width=0.8\columnwidth]{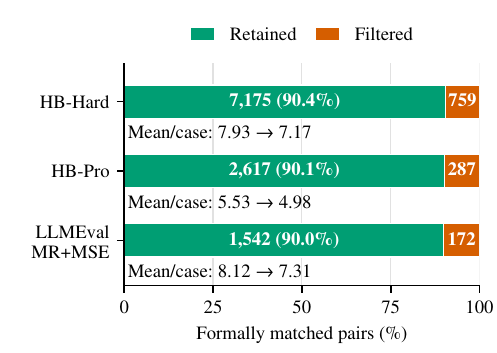}
\caption{Guideline-applicability screening on HB-Hard, HB-Pro, and the LLMEval-Med MR+MSE subset.
A pair is retained if at least one assigned specialty rates it applicable or partially applicable; annotations below the bars show the mean number of guidelines per case before and after screening.}
\label{fig:v2_guideline_applicability}
\end{figure}





\section{Related Work}
\label{sec:related_work}

\paragraph{Generative AI for clinical reasoning.}
Medical language models now span medical question answering~\cite{singhal2023,singhal2025}, domain-specialized models~\cite{li2023,wu2023}, biomedical text generation and mining~\cite{luo2022}, and EHR representation~\cite{yang2022}; broader analyses discuss their benefits and limitations in healthcare~\cite{lee2023,li2024}.
\citet{chen-etal-2025-towards-medical} use reinforcement learning on verifiable medical problems, while \citet{ijcai2024p829} guide clinical reasoning through knowledge seeds.
TriageAgent~\cite{lu-etal-2024-triageagent} combines multi-agent triage with retrieval from the Emergency Severity Index handbook.
Hallucinations and incomplete responses remain concerns~\cite{ji2023,busch2025}. \citet{leung2022} study explainable predictions for trustworthy clinical AI.
\tool{} instead connects source-anchored guideline conditions to patient evidence through an executable intermediate representation, conservative formal matching, and specialty applicability screening.

\paragraph{Trustworthy and neuro-symbolic AI.}
Prompt-based conditioning~\cite{liu2023prompting}, retrieval-augmented generation~\cite{lewis2020rag}, and reinforcement learning from human feedback~\cite{ouyang2022} provide complementary mechanisms for controlling LLM behavior.
Neuro-symbolic systems combine statistical learning and logical structure~\cite{rocktaschel2017,manhaeve2018}, while formal methods have been studied for verification and validation of medical software~\cite{bonfanti2018}.
Examples include model-driven clinical guidance for acute care~\cite{rahmaniheris2016modeldrivendesignclinicalguidance} and verifiable guideline statecharts~\cite{guo2019formalismsupportingdevelopmentverifiably}.
\tool{}'s contribution is to make guideline matching executable and inspectable while using LLMs for bounded semantic judgments, specialty assessment, and response synthesis.

\paragraph{Evaluation of healthcare agents.}
\citet{ijcai2025p1169} present an evaluation taxonomy and highlight causal evaluation as a research direction.
MultifacetEval~\cite{ijcai2024p737} shows that rectification and verification remain difficult even when standard comparisons are strong.
HealthBench~\cite{arora2025} evaluates open-ended healthcare conversations with rubric-based judgments, while LLMEval-Med~\cite{zhang-etal-2025-llmeval} covers complementary medical capabilities.
We evaluate rubric-based answer quality on HealthBench and usability on the medical-reasoning and medical-safety-and-ethics portion of LLMEval-Med, supplemented by component and screening-behavior analyses.

\section{Conclusion}
We presented \tool{}, a formal-constraint–driven framework for refining and aligning LLM-based clinical reasoning with health guidelines. 
Empirical evaluation on public benchmarks demonstrates that formal alignment enhances guideline adherence and reasoning transparency while preserving the flexibility of natural-language interaction.

\label{sec:references_start}
\bibliography{references}

\end{document}